\documentclass[10pt,twocolumn,letterpaper]{article}

 \usepackage[pagenumbers]{cvpr} % To force page numbers, e.g. for an arXiv version

\definecolor{cvprblue}{rgb}{0.21,0.49,0.74}
\usepackage[pagebackref,breaklinks,colorlinks,allcolors=cvprblue]{hyperref}
\usepackage{comment}
\usepackage{multirow}
\usepackage{algorithm}
\usepackage{algorithmic}
\usepackage{float}
\usepackage{stfloats}
\def\paperID{DG-EBF-22} % *** Enter the Paper ID here
\def\confName{CVPR}
\def\confYear{2025}

\title{Task-Level Contrastiveness for Cross-Domain Few-Shot Learning }

\author{
Kristi Topollai \quad Anna Choromanska\\
New York University\\
{\tt\small kt2664@nyu.edu, ac5455@nyu.edu}
}

\begin{document}
\maketitle
\begin{abstract}
Few-shot classification and meta-learning methods typically struggle to generalize across diverse domains, as most approaches focus on a single dataset, failing to transfer knowledge across various seen and unseen domains. Existing solutions often suffer from low accuracy, high computational costs, and rely on restrictive assumptions. In this paper, we introduce the notion of task-level contrastiveness, a novel approach designed to address issues of existing methods. We start by introducing simple ways to define task augmentations, and thereafter define a task-level contrastive loss that encourages unsupervised clustering of task representations. Our method is lightweight and can be easily integrated within existing few-shot/meta-learning algorithms while providing significant benefits. Crucially, it leads to improved generalization and computational efficiency without requiring prior knowledge of task domains. We demonstrate the effectiveness of our approach through different experiments on the MetaDataset benchmark, where it achieves superior performance without additional complexity.
\end{abstract}    
\section{Introduction}
Deep learning has ignited remarkable progress in large scale image classification~\cite{NIPS2012_c399862d} to the point of even surpassing humans. However, deep learning approaches are still not able to efficiently learn new visual concepts and generalize to unseen domains or different tasks. This limits their applicability in the real world scenarios, where the number of data classes is infinite and the classes depend on the task itself. This challenge motivates a few-shot
classification framework~\cite{vinyals2017matching}, where the objective is to learn a model capable of adapting to new
classification tasks given only few labeled samples. The standard approach for this problem is to learn a general case feature extractor, which can then be adapted to new tasks. The meta learning paradigm can be employed to address the few shot learning problem~\cite{vinyals2017matching,finn2017modelagnostic,Sachin2017}, i.e., in this case, episodic training is used to simulate few-shot learning tasks and proper training of the model under both data and task sampling enables it to learn how to generalize on new incoming few-shot learning tasks. 

Most of the traditional meta learning methods rely on a single global model which can learn to generalize on similar tasks, but lacks the representational capacity to accommodate vastly different ones. This results in a performance decrease when tasks come from different domains or even modalities, such as in the case when the few-shot tasks are generated from different data sets \cite{triantafillou2020metadataset}. In this paper we consider this challenging setting and we propose a simple and effective method to improve algorithms operating in this regime. Our approach is based on contrastive learning, a self-supervised learning \cite{balestriero2023cookbook} paradigm where the pretext task is to project similar samples closer in the feature space while pushing dissimilar ones apart, facilitating representation learning without explicit labels. The main contributions of this paper can be summarized as follows

%\textcolor{red}{Write intuitively how do you want to use contrastive learning to fix problems with traditional meta learning methods.}

\begin{itemize}
    \item We extend the contrastive learning framework for the multi task case, where instead of contrasting data representations, as was commonly done before, we instead contrast task representations in an unsupervised manner.
    \item We incorporate our method on the top of multi-domain few-shot and meta learning algorithms and evaluate it on the popular MetaDataset benchmark~\cite{triantafillou2020metadataset}.
    \item We show that our approach, when incorporated into the multi-domain few-shot and meta learning schemes, leads to the following: \textbf{(a)} it improves generalization, \textbf{(b)} it lowers time and memory requirements, \textbf{(c)} it eliminates limiting assumptions, such as knowing the source of each task (i.e. data set of the task, its modality, etc.).
\end{itemize}
\section{Related work}
\label{Related work}
\subsection{Meta learning}
Meta learning methods are commonly categorized into one of the following three families: metric-based, optimization-based, and model-based methods. Metric-based meta-learning algorithms \cite{vinyals2017matching,oreshkin2019tadam,snell2017prototypical} are approaches that focus on learning a metric or distance function during training, allowing a model to quickly generalize its knowledge from a set of base tasks to novel, unseen tasks. These algorithms aim to enable efficient adaptation by capturing similarities and differences between examples, making them particularly useful for tasks like few-shot learning. Optimization-based methods, such as MAML \cite{finn2017modelagnostic}, seek parameters such that the model can quickly adapt  to a new task by running gradient descent starting from these parameters. ANIL \cite{raghu2020rapid} is simillar to MAML, but only adapts the last layer, while R2D2 \cite{bertinetto2019metalearning} and MetaOptNet \cite{lee2019metalearning} replace the last layer with a convex differentiable problem.
Finally, model-based methods learn a model that explicitly adapts to new tasks. The most prominent family of methods is Neural
Processes (NPs) \cite{garnelo2018conditional,gordon2020convolutional} , which encode a support set and estimate
task-specific modulation parameters. All of the above methods adapt well to novel and unseen tasks only if they are similar to the training tasks.

\subsection{Multi-domain few-shot learning}

Handling heterogeneous cases involves constructing models that are aware of the tasks they are designed to tackle. Two mechanisms commonly used to achieve this are: domain expert routing and domain-aware modulation. Methods employing domain-aware modulation \cite{vuorio2019multimodal,yao2019hierarchically,yao2020automated,requeima2020fast,Bateni_2020_CVPR,bateni2022enhancing} aim to adapt a globally shared model using task-specific information. In this setup, a task embedding network is utilized to extract domain-specific details, which are then used to modulate the model parameters. Obtaining an effective task representation is crucial in these methods. Domain expert routing, on the other hand, involves the mechanism of assigning tasks to specific "experts" for processing. These experts can be either different models or model components. Several methods \cite{zhou2020task,10061493,pmlr-v162-jiang22b,10191944,pmlr-v238-lee24b} train the routing component in an unsupervised manner, without prior knowledge of the domain label of each task. Developing a reliable router (or mixer~\footnote{When multiple expert predictions are combined}) is crucial in this scenario.

%domain expert routing refers to the mechanism by which tasks are assigned to specific "experts" for processing. In TSA-MAML~\cite{zhou2020task}, the experts are different models and routing is performed by measuring similarity in parameter space. Similarly \cite{10061493} uses the task's whole gradient and parameter adaptation trajectories for its routing. In MUSE~\cite{10191944}, routing is performed by identifying the expert that would yield the lowest loss for each task. Notably these methods train the routing component in an unsupervised manner, without knowledge of the domain label of each task.
Several cross-domain few shot learning methods employ both aforementioned mechanisms but rely on the knowledge of the domain label of each task. In FLUTE~\cite{triantafillou2021learning} parameters are split into a universal (task agnostic) and task-specific portion. In SUR~\cite{dvornik2020selecting}, they pre-train multiple independent networks (one for each domain) and then, in order to learn in the multi-domain setting, for each task the method identifies the most useful features for this task across domain networks. In URL~\cite{li2021universal,li2022crossdomain} they solve the problem of storing multiple expert networks by
distilling their knowledge to a single set of universal deep representations, while in tri-M~\cite{9710504} each domain is assigned its own domain specific modulation parameters. eTT~\cite{Xu2023ExploringEF} adapts a pre-trained ViT\cite{vit} using an attentive prefix tuning strategy. DIPA~\cite{perera2024discriminative} further improves performance by replacing the standard NCC classifier\cite{requeima2020fast} and introducing a parameter-efficient modulation method. Large foundation models like DINOv2~\cite{oquab2023dinov2} also perform well on few-shot tasks, while Hu et al.~\cite{hu2022pushing} show that a simple pre-training, meta-training, and fine-tuning pipeline can be competitive.
%In TSA-MAML~\cite{zhou2020task}, the experts are different models and routing is performed by measuring similarity in parameter space. Similarly \cite{10061493} uses the task's whole gradient and parameter adaptation trajectories for its routing. In MUSE~\cite{10191944}, routing is performed by identifying the expert that would yield the lowest loss for each task. Notably these methods train the routing component in an unsupervised manner, without knowledge of the mode label of each task.

\subsection{Contrastive learning}
%One of the leading methods in self-supervised learning, contrastive learning, has become a successful paradigm for unsupervised representation learning. Numerous works \cite{hjelm2019learning,oord2019representation,chen2020simple,he2020momentum,radford2021learning,yeh2022decoupled, shidani2024poly} have examined this paradigm before. In the latent space, the contrastive loss seeks to move the anchor sample toward positive samples and away from negative samples. One of the key design choices of the contrastive schemes is the construction of positive and negative samples. The most common approach is to use randomly augmented samples from the anchor sample as positive samples, while all other samples as negative \cite{wu2018unsupervised,chen2020simple,he2020momentum}. In \cite{tian2020makes}, they take this approach one step further and analyze the optimal augmentation strategy in detail. In \cite{robinson2021contrastive,NEURIPS2020_63c3ddcc,sinha2021negative}, they suggest that not all negative samples are informative and empirically analyze the impact of hard negative samples, which they later prioritize when calculating the contrastive loss.

Contrastive learning \cite{hjelm2019learning,oord2019representation,chen2020simple,he2020momentum,radford2021learning,yeh2022decoupled, shidani2024poly} has emerged as a successful approach in self-supervised learning for unsupervised representation learning. The contrastive loss operates in the latent space, aiming to pull the anchor sample towards positive samples while pushing it away from negatives \cite{wu2018unsupervised,chen2020simple,he2020momentum}. A key aspect of contrastive schemes is the construction of positive and negative samples. Typically, randomly augmented samples from the anchor are considered positives, while all others are treated as negatives. Recent work has delved deeper into this, analyzing optimal augmentation strategies~\cite{tian2020makes} and highlighting the importance of informative negative samples, particularly prioritizing hard negatives for contrastive loss calculation \cite{robinson2021contrastive,NEURIPS2020_63c3ddcc,sinha2021negative}.

In the context of meta-learning and cross-domain few-shot learning, \cite{yang2022few} investigates the use of contrastive learning during both pre-training and meta-learning phases to enhance few-shot learning performance. However, their study is limited to single-domain scenarios. In contrast, authors from \cite{gondal2021function}, similar to our approach, leverage contrastive learning to learn a robust representation of the underlying data-generating function (i.e., the data distribution), and validate their method in multi-domain settings. Additionally, in \cite{jang2023unsupervised,poulakakis-daktylidis2024beclr} contrastive learning is applied to improve performance in unsupervised few-shot classification tasks.
\section{Preliminaries}

Few-shot classification is usually formulated as a meta-learning problem. Instead of sampling from the training set, we create a series of learning tasks called episodes, each task
consisting of few labeled examples split into a support set and a query set. Specifically,
in each episode, a small subset of $N$ classes are sampled
from all training classes to construct a support set and a
query set. The support set contains K examples (shots) for each
of the N classes (i.e., N-way K-shot setting) denoted as
$S = \{(X_{1,1}, y_{1,1}),\dots,(X_{K,N}, y_{K,N})\}$, while the query set $Q$ includes different samples from the same $N$ classes. Under the meta learning formalization, the support set is used to adapt the global model and the query set is used to evaluate the adapted model and subsequently update the global model. This learning setup can be summarized as 
\begin{equation}
\min_{\theta} \sum_{i=1}^{|T|}\mathcal{L}(Q_{i}, \mathcal{A}(\theta, S_{i})),
\end{equation}
where $|T|$ is the number of training tasks/episodes, $\mathcal{L}$ some loss function and $\mathcal{A}$ the adaptation operator, which given a parameter set $\theta$ and a support set $S$, outputs a set of task adapted parameters $\hat{\theta} = \mathcal{A}(\theta, S)$. In MAML\cite{finn2017modelagnostic}, the operator $\mathcal{A}(\theta, S)$ is a standard gradient descent, $\hat{\theta} = \theta - \alpha\nabla_{\theta}\mathcal{L}(\theta, S)$. In the multi-domain case, the support and query sets $T = (S,Q)\sim D $ are not sampled from a single task distribution $D$, but may come from one of several such distributions. Lets call the distribution over potential domains $\mathbb{D}$, then the goal is to minimize the expected loss over all domains:
\begin{equation}\min_{\theta}\mathbb{E}_{D\sim \mathbb{D}}[\mathbb{E}_{(S,Q)\sim D}[\mathcal{L}(Q, \mathcal{A}(\theta, S))]].\end{equation}
Naturally, the adaptation $\mathcal{A}$ in this case should be more sophisticated, so that it considers the domain information itself in addition to the support set samples. To this end, our focus will be on how to aid the unsupervised domain identification and domain specific adaptation of a task. Our method is based on contrastive learning and in particular SimCLR~\cite{chen2020simple} where authors proposed a simple framework for contrastive learning of visual representations. The process is simple, first, a random minibatch of $N$ samples is drawn and for each sample two different data augmentations are applied, resulting in $2 N$ augmented samples in total. This is formalized as follows:
$$
\mathbf{x}_1=t(\mathbf{x}), \quad \mathbf{x}_2=t^{\prime}(\mathbf{x}), \quad t, t^{\prime} \sim \mathcal{T},
$$
where $t$ and $t^{\prime}$ denote two separate data augmentation operators that are sampled from the same family of augmentations $\mathcal{T}$. $(x_1,x_2)$ is one positive pair and the rest $2(N-1)$ data points are treated as negative samples. The contrastive learning loss is based on the NT-Xent loss, using cosine similarity $\operatorname{sim}(.,.)$ on some projection $g$ of the representations $f_{\theta}\left(x\right)$.
$$
\mathbf{z}_1=g\left(f_{\theta}\left(\mathbf{x}_1\right)\right), \quad \mathbf{z}_2=g\left(f_{\theta}\left(\mathbf{x}_2\right)\right).
$$
\begin{equation}
\label{eq:lcon2}
\mathcal{L}_{\text {SimCLR }}^{(i, j)} =-\log \frac{\exp \left(\operatorname{sim}\left(\mathbf{z}_i, \mathbf{z}_j\right) / \tau\right)}{\sum_{k=1}^{2 N} \mathbb{I}_{[k \neq i]} \exp \left(\operatorname{sim}\left(\mathbf{z}_i, \mathbf{z}_k\right) / \tau\right)}\end{equation}
%where $\mathbb{I}_{[k \neq i]}$ is an indicator function and $\tau$ is the temperature hyperparameter.

\section{Task-level contrastive learning}

\label{method}
We propose using task-level contrastiveness to improve the performance of few-shot/meta learning schemes. Our approach can be used on top of most algorithms in the literature and as we will show in the experimental section, provides significant performance boosts. In a manner similar to traditional contrastive learning, the objective is to bring representations of similar tasks close to each other and push away dissimilar ones.

\subsection{Task representations}
\label{tr}
There are several ways to define a task representation, most modulation based few-shot and meta learning algorithms explicitly define a task encoding network $g_{\phi}$ parametrized by $\phi$ to construct task embeddings, which are then used to adapt the model to the target task. Such an encoder should include 1) high representational capacity, and 2) permutational invariance to its inputs. A commonly used encoder that fits the requirements is based on DeepSets ~\cite{zaheer2018deep} and uses convolutional blocks to produce rich representations, while the average pooling operation in both spatial and batch dimension maintains the permutation invariance. The operation of the encoding given a support set $S$ is:
\begin{equation}z = g_{\phi}(S)\end{equation}
Another possibility is to use the
feature extractor $f_{\theta}$ itself and use the individual sample embeddings to construct task embeddings. In addition we can also consider either the pre or post adaptation features, which would allow for label-aware representations. To obtain a permutation invariant task embedding we can aggregate the individual representations with a permutation invariant operator, such as for example mean, or max/min, etc. In case of using the mean, the task embedding is given as:
\begin{equation}
z = \frac{1}{N}\sum_{(x_{i},y_{i})\in S}f_{\theta}(x_{i}).
\end{equation}

\subsection{Task augmentations}
\label{sec:ta}
\begin{figure*}
\begin{center}
\centerline{\includegraphics[width=0.9\textwidth]{
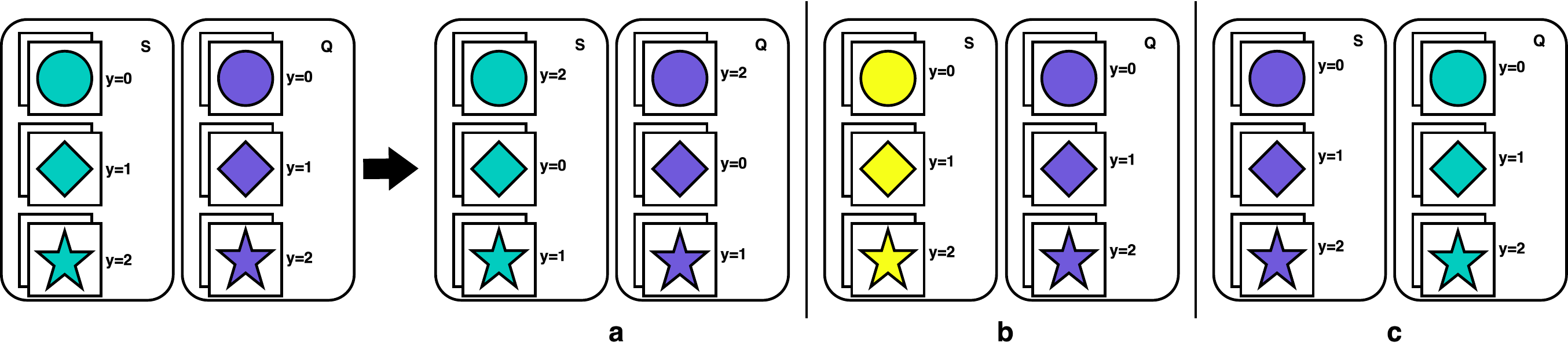}}
\caption{The 3 proposed task augmentation strategies (a) relabeling, (b) instance augmentation , and (c) mixing.}
\label{fig:schematics}
\end{center}
\vspace{-0.4in}
\end{figure*}

Since our main assumption is the lack of domain labels for each task, using tasks from the same domain as positive pairings to define the contrastive loss is not an option. We can however approach this problem similarly to SimCLR by treating two augmented views of the same task as a positive pair and every other task in the batch as negative. Luckily defining task augmentations for contrastive purposes is easier than in the case of instance-level augmentations. We will consider three simple straightforward approaches, illustrated in Figure~\ref{fig:schematics}, different from general task-level augmentations found in non contrastive settings such as in \cite{ni2021data}:

     \textbf{Relabeling} It is exclusive to few shot classification. The idea is to relabel the samples in the task. Define $P$ as the permutation operator which assigns a new label to all samples in the same class, then a positive task pair, given a task $T = (S,Q)$ and two permutation operators $P,P'$, is defined as
     \begin{align}
         T_{1} &= P(S,Q) = \{(X,P(y))\}_{(X,y)\in S\cup Q}\\
         T_{2} &= P^{\prime}(S,Q) = \{(X,P^{\prime}(y))\}_{(X,y)\in S\cup Q}
     \end{align}
     \textbf{Instance augmentation} This task augmentation method is a direct extension of sample-level augmentation in SimCLR. Let $t_{1},t_{2} \sim \mathcal{T}$ be data augmentations, then the task pair, given a task $T = (S,Q)$, is 
     \begin{align}
         T_{1} &= (t_{1}(S),Q) = (\{(t_{1}(X),y)\}_{(X,y)\in S},Q)\\
         T_{2} &= (t_{2}(S),Q) = (\{(t_{2}(X),y)\}_{(X,y)\in S},Q).
     \end{align}
    Note that while not necessary, augmentations can be applied on the query samples as well.
    
     \textbf{Mixing}
    The third task augmentation that we discuss, is mixing of the support and query sets. Intuitively, this augmentation method is the most powerful, as the mutual information between the two task views is smaller compared to the above two augmentation strategies~\cite{tian2020makes}. First we define the number of samples $M$ that we want to substitute from the support set. The most powerful augmentation, meaning the one for which the two task embeddings come from the most distinct support sets, would be the one for which $M$ is set to $M = \min(|S|,|Q|)$. Note that in order to maintain the class balancedness of each set, we substitute $M/K$ samples from each class, where $K$ is the shots of the task. Define the operator $\mathcal{M}_{M}(\cdot,\cdot)$ that performs this exact mixing and that is drawn from a distribution of such operators, i.e., $\mathcal{M}_{M}(\cdot,\cdot) \sim \mathbb{M}_{M}$. For each class in the support set, it substitutes $M/K$ samples from each class with $M/K$ samples from the query set, which comes from the same class. Given $\mathcal{M}_{M}, \mathcal{M}_{M}'\sim \mathbb{M}_{M}$ the task pair, given a task $T = (S,Q)$, is then defined as
    \begin{align}
         T_{1} &= \mathcal{M}_{M}(T) = \mathcal{M}_{M}(S,Q)\\
         T_{2} &= \mathcal{M}_{M}'(T) = \mathcal{M}_{M}'(S,Q).
     \end{align}

\subsection{Enhancing Domain Aware Modulation and Domain Expert Routing Mechanisms}
As mentioned in the related work section, we identify two mechanisms utilized by methods operating in this regime, domain-aware modulation and domain expert routing. Both mechanisms can be enhanced within our framework by:
\textbf{(a)} Contrasting task representations of different domains to encourage domain-aware modulation.
\textbf{(b)} Contrasting task representations to further refine the implicit clustering of task representations and, consequently, the domain routing mechanism.
\textbf{(c)} Employing contrastive clustering to enable unsupervised domain routing and eliminate the domain label assumption.

\section{Three case studies}

For proof of concept, we selected three methods (MMAML, TSA-MAML, Tri-M) due to their simplicity, facilitating a clear demonstration of our task-level contrastive approach. Specifically:\textbf{(a)} MMAML: To demonstrate how task-aware modulation can be improved. \textbf{(b)} TSA-MAML: To illustrate improvements in unsupervised domain routing. \textbf{(c)} Tri-M: To showcase how to enable end-to-end unsupervised domain routing by removing the domain label assumption.

Given that most other methods in this field utilize one or both of these mechanisms, they can be enhanced under our general framework by:
\textbf{(a)} Identifying which of the aforementioned mechanisms are at play.
\textbf{(b)} Enhancing the mechanism through the utilization of task-level contrastiveness.

%It is noteworthy that while our method is highly general, other methods employing these mechanisms can also be improved under our framework, albeit in a slightly more ad-hoc manner (e.g., TSA-ANIL). This approach can yield additional benefits such as reduced computational requirements. Additionally, we emphasize that our goal is to introduce an additional tool that can be seamlessly integrated into existing methods, rather than proposing a novel standalone method.

\subsection{Enhancing domain aware modulation}
\begin{figure*}
\begin{center}
\centerline{\includegraphics[width=0.8\textwidth]{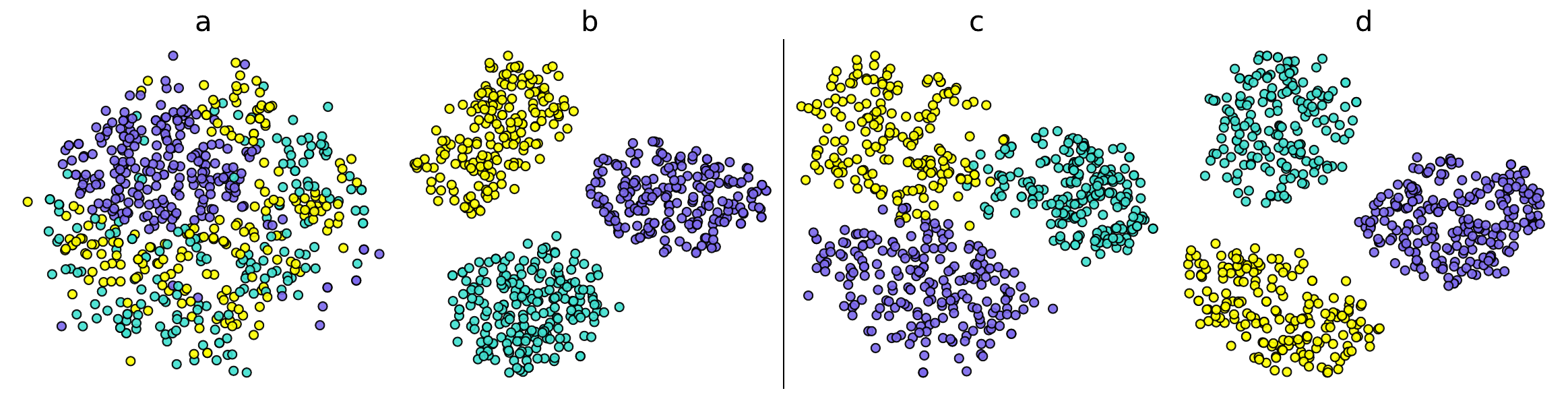}}
\caption{5 way 1 shot (a,b) and 5 shot (c,d) experiments on fungi, aircraft, birds datasets. We project the adapted parameters for MAML in (a,c) and the task representations for MAML with contrastive loss in (b,d). Unlike task representations, parameters are not a reliable router for TSA-MAML. }
\label{fig:cluster}
\end{center}
\vspace{-0.4in}
\end{figure*}
\subsubsection{MMAML}

The first algorithm we consider is MMAML \cite{vuorio2019multimodal}, one of the first approaches handling heterogeneous task distributions whose task dependent parameter modulation approach is fundamental and used in a plethora of powerful methods \cite{bateni2022enhancing,li2022crossdomain}. MMAML and other modulation based methods consist of two complementary neural networks. A modulation network $g_{\phi}$ produces a task embedding $z$ , which is used
to generate parameters $\tau_{i}$ that modulate the
task network. The task network $f_{\theta}$ adapts those modulated
parameters (MAML-like gradient based adaptation) to fit to the target task. This extra step of modulation, ensures that the initialization is more domain specific. The MMAML training can be summarized by the following optimization problem:
\begin{equation}
\begin{split}
    \min_{\theta} & \quad \sum_{i=1}^{T}\mathcal{L}(Q_{i}, \mathcal{A}_{\phi}(\theta, S_{i}))\\
\text{s.t.} & \quad \mathcal{A}_{\phi}(\theta, S_{i}) = \hat{\theta}_{0} - a\sum_{t=1}^{T}\nabla_{\theta}\mathcal{L}(Q_{i}, \hat{\theta}_{t})
\end{split}\end{equation}
and the pre adaptation parameters $\hat{\theta}_{0}$, are obtained by the following task specific modulation
$$\hat{\theta}_{0} = \tau\odot\theta ,\quad \tau = h_{w}(g_{\phi}(S_{i})).$$
The modulation parameters $\tau$ are produced by a simple network which takes as an input a task embedding $g_{\phi}(S_{i})$. The modulation operation is performed by the FiLM layer \cite{perez2017film}.  Minimizing this loss produces reasonable task embeddings, which in turn produce good parameter modulations. 

\subsubsection{Contrastive MMAML}
\label{cmmaml}
We build on the top of MMAML and further encourage good task embeddings, which capture the properties of individual domains, by adding a task-level contrastive term and thus changing the MMAML optimization problem into:
\begin{equation}\min_{\theta,\phi}( \lambda\mathcal{L}_{\text{con}}+\sum_{i=1}^{T}\mathcal{L}(Q_{i}, \mathcal{A}_{\phi}(\theta, S_{i})).\end{equation}
If $B = (S_{1},Q_{1}),\dots,(S_{N},Q_{N})$ is the task batch, we apply one of the task augmentations from Section~\ref{sec:ta} on each task to obtain positive pairs $(S_{2k-1}',Q_{2k-1}'),(S_{2k}',Q_{2k}')$ for $k=1,\dots N$.  The augmented batch is then $B' = (S_{1}',Q_{1}'),\dots,(S_{2N}',Q_{2N}')$. We can use the MMAML loss on this batch and in addition define the contrastive loss $\mathcal{L}_{\text{con}}$ using the NT-Xent loss~\ref{eq:lcon2}  $L(i,j)$ on the set of task embeddings $\{z_{i}\}_{i=1}^{2N} = \{g_{\phi}(S_{i}')\}_{i=1}^{2N}$ as follows:
\begin{equation}
\label{eq:lcon1}
\mathcal{L}_{\text{con}} = \frac{1}{2N}\sum_{k=1}^{2N}(L(2k-1,2k)+L(2k,2k-1)).\end{equation}

\subsection{Improving unsupervised domain routing }
\subsubsection{TSA-MAML}

In TSA-MAML~\cite{zhou2020task}, we assume a MAML model pre-trained on tasks from all domains. However, instead of relying on this shared global initialization, it obtains domain specific initializations, by first sampling many tasks from all modalities and then clustering their adapted parameters. To fine-tune those domain specific initializations, for each task, a trial adaptation is performed to identify its domain. This domain corresponds to the cluster, whose center lies closest to its trial adapted parameters. One obvious drawback of this method is the need for a trial adaptation, which adds a significant time overhead. Another drawback is the need to store multiple domain specific networks. Furthermore, this method assumes that this parameter-space based routing is reliable. While this is possible in some cases, it also heavily depends on the experiment itself (Fig:~\ref{fig:cluster}(a,c)).

\begin{table*}[t]
\caption{Task augmentation quality evaluation, MMAML and TSA-MAML are run on 3 datasets, 5 ways, 1 shot}
\vspace{-0.2in}
\label{sample-table}
\begin{center}
\begin{tabular}{lllll}
\cline{2-5}
                                    & \textbf{MetaDataset} &                            & \textbf{MMAML } & \textbf{TSA-MAML } \\ \hline
\textbf{Task Augmentation strategy} & \textbf{DB}          & \textbf{Linear Classifier} & \textbf{Accuracy}     & \textbf{Accuracy}        \\ \hline
No Augment                          & 1.53                 & 86.6\%                     & 56.6\%                & 51.5\%                   \\
Relabeling                          & 1.79                 & 80.8\%                     & 56.5\%                & 53.2\%                   \\
Mixing                              & \textbf{0.25}        & \textbf{99.1\%}            & \textbf{57.6\%}       & \textbf{53.6\%}          \\
Image Augment                       & 0.62                 & 93.4\%                     & 56.8\%                & 53.5\%                   \\
Mixing+Relabeling                   & \textbf{0.24}        & \textbf{99.1\%}            & \textbf{57.6\%}       & \textbf{53.7\%}          \\
Mixing+Image Augment                & 0.51                 & 95.8\%                     & 56.9\%                & 53.4\%                   \\
Relabeling+Mixing+Image Augment     & 0.50                 & 95.8\%                     & 56.8\%                & 53.5\%                   \\ \hline
\end{tabular}
\end{center}
\label{table:augs}
\vspace{-0.3in}
\end{table*}

\subsubsection{Contrastive TSA-MAML}
Building on the observation that tasks are not nicely separable in the parameter space, we instead rely on task representations, which under our contrastive framework can form well-separated clusters (see Fig: \ref{fig:cluster}, (b,d)). Instead of using the optimal parameters to perform the routing, we instead use the task representations. We suggest the feature aggregation we introduced in Section~\ref{tr}. The main benefit of relying on the features, is that, we can easily encourage contrastive features and in turn task representations by incorporating a contrastive loss. We can therefore improve TSA-MAML by changing the vanilla MAML initalization, to MAML trained with contrastive loss~\ref{eq:lcon1}. If $f_{\theta}$ is our feature extractor and $w$ are the parameters of the linear classifier layer, then the contrastive MAML loss when using the post or pre-adaptation features
$\{z_{i}\}_{i=1}^{2N} = \{\frac{1}{kN}\sum_{x\in S_{i}}f_{\mathcal{A}(\theta)}(x)\}_{i=1}^{2N}$ is:
\begin{equation}
\begin{split}
    \min_{w,\theta} & \quad (\lambda\mathcal{L}_{\text{con}}+\sum_{i=1}^{T}\mathcal{L}(Q_{i}, \mathcal{A}(w,\theta, S_{i})))\\
\text{s.t.} & \quad \mathcal{A}(w,\theta, S_{i}) = (w,\theta) - a\nabla\mathcal{L}(S_{i}, (w,\theta))
\end{split}\end{equation}

Our contrastive approach offers another advantage: a reduction in both space and time requirements. Thanks to the reliable expert routing mechanism within our task contrastive framework, we gain greater flexibility in defining the networks for our experts. Instead of each expert constituting an entire network, they can be defined as subsets of parameters. One option we propose is inspired by the findings of Raghu et al~\cite{raghu2020rapid}. Their work demonstrated that by omitting the inner loop adaptation of the entire feature extractor and only adapting the classification head, we achieve nearly identical results. Therefore, in our case, each expert would consist of a separate classification head. Another possible option, particularly advantageous for large models, involves bias tuning~\cite{10.5555/3495724.3496671,ben-zaken-etal-2022-bitfit}. In this scenario, the experts would be composed of different sets of biases.

The computational advantages are two. Firstly, our domain-specific experts are smaller subsets of parameters, requiring much less memory to store compared to a full vision model. Secondly, we eliminate the need for trial adaptations to perform routing; a single forward pass is sufficient to identify the task domain.
%The fact that we are using pre-adaptation features, makes the contrastive training all the more important and enables the usage of the ANIL instead of MAML for the TSA backbone.

\begin{table}[h!]
\centering
\caption{Computational requirements, $N$ is the size of the main model, $n$ the size of the domain expert models ($n<N$) and $M$ the number of domain experts}
\begin{tabular}{lll}
\hline
                                  & \textbf{\begin{tabular}[c]{@{}l@{}}Parameters \\ in memory\end{tabular}} & \textbf{\begin{tabular}[c]{@{}l@{}}Trial \\ Adaptation\end{tabular}} \\ \hline
\textbf{TSA-MAML}                 & N+MN                                                                     & Yes                                                                  \\
\textbf{our c-TSA-MAML} & N+Mn                                                                     & No                                                                   \\ \hline
\end{tabular}
\label{table:computations}
\vspace{-0.2in}
\end{table}

\subsection{End-to-end unsupervised domain expert routing}
A common assumption in few-shot learning methods that limits their real-world applicability is the requirement for domain labels, indicating the source of each task. This knowledge is crucial for training distinct expert networks, as seen in \cite{li2021universal}, or for training a domain classifier, as demonstrated in \cite{triantafillou2021learning,9710504}. Notably, in \cite{9710504}, the domain classifier must be trained concurrently with the rest of the architecture, making non-contrastive deep clustering methods impractical~\cite{caron2018deep}. In this section, we introduce task contrastive clustering and its application in Tri-M~\cite{9710504}. In our experimental section, we also present results demonstrating optimal performance in unsupervised task domain classification on MetaDataset. These results serve as evidence of the applicability of our method in FLUTE~\cite{triantafillou2021learning} and URL~\cite{li2021universal}, both of which rely on a domain-router trained with supervision.

\subsubsection{Task Contrastive Clustering}

\begin{table*}[t]
\caption{The performance (classification accuracy) for the MMAML and its contrastive counterpart}
\small
\vspace{-0.2in}
\label{table:mmaml}
\begin{center}
\begin{tabular}{ll|llll|llll}
\cline{3-10}
                          &                    & \multicolumn{4}{c|}{\textbf{1 Shot}}                                                                                                         & \multicolumn{4}{c}{\textbf{5 Shot}}                                                                                                         \\ \hline
                          &                    & \textbf{MAML}        & \textbf{MMAML}       & \multicolumn{1}{l|}{\textbf{\begin{tabular}[c]{@{}l@{}}contrastive\\ MMAML\end{tabular}}} & \textbf{\begin{tabular}[c]{@{}l@{}}supervised\\ MMAML\end{tabular}} & \textbf{MAML}        & \textbf{MMAML}       & \multicolumn{1}{l|}{\textbf{\begin{tabular}[c]{@{}l@{}}contrastive\\ MMAML\end{tabular}}} & \textbf{\begin{tabular}[c]{@{}l@{}}supervised\\ MMAML\end{tabular}} \\ \hline
\multirow{4}{*}{\rotatebox{90}{\textbf{3 datasets}}}     & \textbf{Omniglot}     & $86.1_{\pm 1.1}$     & $92.6_{\pm 1.2}$     & \multicolumn{1}{l|}{$93.8_{\pm 0.8}$}                                                     & $93.8_{\pm 0.5}$                                                    & $88.9_{\pm 0.9}$     & $91.8_{\pm 0.9}$     & \multicolumn{1}{l|}{$93.1_{\pm 0.7}$}                                                     & $93.1_{\pm 0.6}$                                                    \\
                          & \textbf{MiniImage}  & $42.4_{\pm 1.3}$     & $42.0_{\pm 0.9}$     & \multicolumn{1}{l|}{$42.9_{\pm 0.7}$}                                                     & $43.4_{\pm 0.5}$                                                    & $52.7_{\pm 0.6}$     & $54.7_{\pm 0.7}$     & \multicolumn{1}{l|}{$56.6_{\pm 0.5}$}                                                     & $56.8_{\pm 0.4}$                                                    \\
                          & \textbf{FC100}       & $33.6_{\pm 0.7}$     & $35.2_{\pm 1.0}$     & \multicolumn{1}{l|}{$36.1_{\pm 0.7}$}                                                     & $36.3_{\pm 0.4}$                                                    & $47.8_{\pm 0.5}$     & $44.0_{\pm 0.6}$     & \multicolumn{1}{l|}{$47.2_{\pm 0.4}$}                                                     & $47.3_{\pm 0.4}$                                                    \\
                          & \textbf{Avg}         & $54.0_{\pm 1.1}$     & $56.6_{\pm 1.0}$     & \multicolumn{1}{l|}{$\mathbf{57.6_{\pm 0.7}}$}                                            & $57.8_{\pm 0.5}$                                                    & $63.1_{\pm 0.7}$     & $63.5_{\pm 0.7}$     & \multicolumn{1}{l|}{$\mathbf{65.6_{\pm 0.5}}$}                                            & $65.7_{\pm 0.4}$                                                    \\ \hline
\multirow{6}{*}{\rotatebox{90}{\textbf{5 datasets}}}     & \textbf{Omniglot}     & $84.4_{\pm 1.0}$     & $90.7_{\pm 1.2}$     & \multicolumn{1}{l|}{$91.0_{\pm 0.6}$}                                                     & $91.4_{\pm 0.5}$                                                    & $89.4_{\pm 0.9}$     & $93.5_{\pm 1.0}$     & \multicolumn{1}{l|}{$91.3_{\pm 0.8}$}                                                     & $91.2_{\pm 0.8}$                                                    \\
                          & \textbf{MiniImage}  & $39.6_{\pm 1.3}$     & $40.0_{\pm 1.1}$     & \multicolumn{1}{l|}{$42.4_{\pm 0.7}$}                                                     & $42.7_{\pm 0.6}$                                                    & $51.3_{\pm 0.7}$     & $53.3_{\pm 0.8}$     & \multicolumn{1}{l|}{$55.6_{\pm 0.6}$}                                                     & $56.0_{\pm 0.5}$                                                    \\
                          & \textbf{FC100}       & $34.1_{\pm 0.9}$     & $34.9_{\pm 0.9}$     & \multicolumn{1}{l|}{$34.7_{\pm 1.0}$}                                                     & $34.7_{\pm 0.8}$                                                    & $43.4_{\pm 0.7}$     & $43.0_{\pm 0.7}$     & \multicolumn{1}{l|}{$48.2_{\pm 0.5}$}                                                     & $48.2_{\pm 0.5}$                                                    \\
                          & \textbf{Birds}       & $37.9_{\pm 0.9}$     & $43.0_{\pm 1.0}$     & \multicolumn{1}{l|}{$43.8_{\pm 0.8}$}                                                     & $43.7_{\pm 0.6}$                                                    & $47.9_{\pm 0.5}$     & $53.2_{\pm 0.6}$     & \multicolumn{1}{l|}{$60.0_{\pm 0.4}$}                                                     & $60.3_{\pm 0.4}$                                                    \\
                          & \textbf{Aircraft}    & $44.6_{\pm 1.3}$     & $43.1_{\pm 1.2}$     & \multicolumn{1}{l|}{$44.5_{\pm 0.7}$}                                                     & $45.0_{\pm 0.5}$                                                    & $56.6_{\pm 1.0}$     & $54.1_{\pm 1.0}$     & \multicolumn{1}{l|}{$53.8_{\pm 0.8}$}                                                     & $54.0_{\pm 0.6}$                                                    \\
                          & \textbf{Avg}         & $48.1_{\pm 1.1}$     & $50.3_{\pm 1.1}$     & \multicolumn{1}{l|}{$\mathbf{51.3_{\pm 0.8}}$}                                            & $51.5_{\pm 0.6}$                                                    & $57.7_{\pm 0.8}$     & $59.4_{\pm 0.8}$     & \multicolumn{1}{l|}{$\mathbf{61.8_{\pm 0.6}}$}                                            & $61.9_{\pm 0.5}$                                                    \\ \hline
\end{tabular}
\end{center}
\vspace{-0.2in}
\end{table*}

\begin{table*}[t]

\caption{The performance (classification accuracy) for the TSA-MAML and its contrastive counterpart}
\small
\vspace{-0.2in}
\label{table:tsamaml}
\begin{center}
\begin{tabular}{ll|lllllll}
\hline
                   &                    & \textbf{MAML}    & \textbf{ANIL}    & \textbf{TSA-MAML} & \textbf{TSA-ANIL} & \textbf{\begin{tabular}[c]{@{}l@{}}contrastive\\ TSA-MAML\end{tabular}} & \textbf{\begin{tabular}[c]{@{}l@{}}contrastive\\ TSA-ANIL\end{tabular}} & \textbf{\begin{tabular}[c]{@{}l@{}}supervised\\ TSA-MAML\end{tabular}} \\ \hline
\multirow{4}{*}{\rotatebox{90}{\textbf{3 datasets}}} & \textbf{Fungi}     & $40.9_{\pm 1.0}$ & $40.8_{\pm 1.1}$ & $41.1_{\pm 1.3}$  & $41.0_{\pm 1.2}$  & $42.7_{\pm 0.9}$ & $42.4_{\pm 0.8}$ & $42.6_{\pm 0.8}$ \\
                   & \textbf{Birds}     & $52.9_{\pm 0.9}$ & $52.7_{\pm 1.0}$ & $53.2_{\pm 1.2}$  & $52.8_{\pm 1.1}$  & $54.4_{\pm 0.8}$ & $53.6_{\pm 0.8}$ & $54.6_{\pm 0.7}$ \\
                   & \textbf{Aircraft}  & $60.8_{\pm 1.2}$ & $60.8_{\pm 1.3}$ & $61.0_{\pm 1.3}$  & $61.1_{\pm 1.3}$  & $63.5_{\pm 0.8}$ & $62.7_{\pm 0.9}$ & $63.8_{\pm 0.8}$ \\
                   & \textbf{Avg}       & $51.5_{\pm 1.0}$ & $51.4_{\pm 1.1}$ & $51.8_{\pm 1.3}$  & $51.6_{\pm 1.2}$  & $\mathbf{53.6_{\pm 0.8}}$ & $52.9_{\pm 0.8}$ & $53.7_{\pm 0.8}$ \\ \hline
\multirow{6}{*}{\rotatebox{90}{\textbf{5 datasets}}} & \textbf{Omniglot}  & $93.2_{\pm 1.0}$ & $94.0_{\pm 1.3}$ & $93.8_{\pm 1.1}$  & $93.8_{\pm 1.3}$  & $94.5_{\pm 0.9}$ & $94.2_{\pm 1.0}$ & $94.4_{\pm 0.8}$ \\
                   & \textbf{MiniImage} & $39.9_{\pm 1.1}$ & $39.1_{\pm 1.2}$ & $40.0_{\pm 1.2}$  & $39.7_{\pm 1.2}$  & $44.0_{\pm 0.8}$ & $42.8_{\pm 0.9}$ & $44.3_{\pm 0.7}$ \\
                   & \textbf{Fungi}     & $40.7_{\pm 0.8}$ & $40.9_{\pm 1.0}$ & $40.6_{\pm 0.9}$  & $40.9_{\pm 1.1}$  & $42.1_{\pm 0.7}$ & $41.7_{\pm 0.9}$ & $42.1_{\pm 0.8}$ \\
                   & \textbf{Birds}     & $53.1_{\pm 0.9}$ & $52.5_{\pm 1.0}$ & $53.4_{\pm 1.0}$  & $52.8_{\pm 1.0}$  & $55.1_{\pm 0.8}$ & $53.9_{\pm 0.8}$ & $54.9_{\pm 0.8}$ \\
                   & \textbf{Aircraft}  & $60.5_{\pm 1.1}$ & $59.9_{\pm 1.1}$ & $60.7_{\pm 1.2}$  & $60.5_{\pm 1.1}$  & $61.9_{\pm 0.9}$ & $61.9_{\pm 1.0}$ & $62.5_{\pm 0.8}$ \\
                   & \textbf{Avg}       & $57.5_{\pm 1.0}$ & $57.3_{\pm 1.1}$ & $57.7_{\pm 1.1}$  & $57.5_{\pm 1.1}$  & $\mathbf{59.5_{\pm 0.8}}$ & $58.9_{\pm 0.9}$ & $59.6_{\pm 0.8}$ \\ \hline
\end{tabular}
\end{center}
\vspace{-0.3in}
\end{table*}

Contrastive Clustering~\cite{li2020contrastive} first learns the feature
matrix (the matrix of instance representations). After that,
the instance- and cluster-level contrastive learning are conducted in the row and column space of the feature matrix by
gathering the positive pairs and scattering the negatives. By
considering the instance- and cluster-level similarity under
their dual contrastive learning framework, Contrastive Clustering is able to simultaneously learn discriminative features and perform clustering in an online and end-to-end manner. This is possible by optimizing the loss:
\begin{equation}\mathcal{L}_{\text{con}} = \mathcal{L}_{\text{con}}^{\text{task}}+\mathcal{L}_{\text{con}}^{\text{clu}}-\mathcal{H},\end{equation}
where the term $\mathcal{L}_{\text{con}}^{\text{task}}$ is the task embedding contrastive loss we introduced before, while $\mathcal{L}_{\text{con}}^{\text{clu}}$ is again the contrastive loss, applied on the columns of the task embedding matrix after being projected on a space with dimensionality equal to the number of domains. This projection acts as a soft label of the domain. To avoid the trivial
solution in which most instances are assigned to the same cluster, Contrastive Clustering also includes the loss term $\mathcal{H}$ that is the entropy of cluster assignment probabilities. Using our task-level contrastive loss, leads to task-level contrastive clustering.

\subsubsection{Tri-M Algorithm}
The tri-M algorithm is based on task specific modulation, the main difference being, the usage of two sets of modulation parameters. The
domain-specific set and the domain-cooperative set, which
work complementarily to explore both the intra-domain and
inter-domain information. The modulation is then performed again by FiLM layers. The authors treat the problem of choosing the appropriate modulation parameters as a domain classification problem, where a task encoder is trained using the domain labels.  By substituting the supervised cross entropy loss of the task encoder of Tri-M with our Task Contrastive Clustering loss, we are able to overcome the need for domain labels. This gives rise to our Contrastive Tri-M.

% Concretely, the domain-specific set describes each domain independently to prevent negative interference among distant domains, e.g., ImageNet and Omniglot. The domain-cooperative set captures the inter-domain relations to guide beneficial information sharing among relevant domains to enrich their own domain specific representations.

\section{Experiments}

\label{experiments}

\subsection{Ablation study on task augmentations}
We first conduct a simple ablation study to choose the most promising task augmentation for both routing and modulation based methods. In Table \ref{table:augs}, we first evaluate the quality of the clustering achieved. We present the Davies-Bouldin index and the linear classification accuracy, where we use task embeddings as features and their domains as labels. We also present the accuracy for contrastive MMAML and TSA-MAML. Notably, mixing outperforms other methods. Relabeling, while beneficial for TSA-MAML with post-adaptation embeddings, performs poorly with MMAML (and Tri-M) due to collapsed representations. The collapse occurs because the label is only included in the input of the task encoder, which does not ensure sufficiently distinct embeddings for a positive pair of tasks. Instance augmentation demonstrates strong performance. However, we hypothesize that the composition of augmentations in SimCLR may not be optimal for our own datasets, resulting in a suboptimal feature extractor. Lastly, achieving optimal performance on the unsupervised task domain classification in MetaDataset serves as direct evidence of our method's applicability in FLUTE~\cite{triantafillou2021learning} and URL~\cite{li2021universal}, both of which rely on a domain-router trained in a supervised manner.

\subsection{Experimental results}
For all three methods and their contrastive counterparts we use the exact same experimental setting as in the respective papers. We refer to the original papers and our Supplement for the detailed experimental settings. We also refer to our Supplement for more experimental results, in particular 5-shot experiments for the TSA-MAML family of methods. In the case of MMAML, we see in Table \ref{table:mmaml} that adding contrastiveness results in better performance. This is because it improves the ability to distinguish between different domains and in turn results in better modulation parameters. TSA-MAML, Table~\ref{table:tsamaml}, also benefits from contrastiveness due to proper domain assignment, while, contrastive TSA-ANIL, which is our proposed method, where experts are different classification heads instead of full networks, exhibits favorable performance compared to TSA-MAML at a fraction of the computational requirements of TSA-MAML.
\begin{table*}[t!]
\caption{Computational benefits of (a) TSA-MAML variants on 3 datasets 5 ways 1 shot and (b) URL on Metadataset}
\small
\vspace{-0.2in}
\label{sample-table}
\begin{center}
\begin{tabular}{@{}llllllll@{}}
\cmidrule(r){1-4} \cmidrule(l){6-8}
\textbf{Frozen Layers}  & \textbf{Avg Acc} & \textbf{\#Params} & \textbf{Trial Adaptation} &  &                                   & \textbf{Avg Acc} & \textbf{\#Nets} \\ \cmidrule(r){1-4} \cmidrule(l){6-8} 
\textbf{0 (TSA-MAML)}   & 51.7             & 484372/100\%                              & Yes                       &  &  \textbf{Single task networks}     &   76.4      &                  8                \\
\textbf{0 (c-TSA-MAML)} & 53.6             & 484372/100\%                              & No                        &  & \textbf{Random grouping}          &  76.3       &              3                    \\
\textbf{3 (c-TSA-ANIL)}          & 52.9             & 256276/52.9\%                             & No                        &  & \textbf{Task similarity grouping} & 76.6        &              3                \\ \cmidrule(r){1-4} \cmidrule(l){6-8} 
\end{tabular}
\end{center}
\label{table:compbenefits}
\vspace{-0.3in}
\end{table*}

Both non-contrastive MMAML and TSA-MAML rely on implicit clustering of task representations and parameters, which cannot be reliably achieved without an explicit loss. Consequently, non-contrastive experiments exhibit higher standard deviation in the final accuracy. Proper discrimination between domains is further demonstrated in Figure \ref{fig:cluster} (b, d). It can be observed that relying on clustering in the parameter space may cause all centers to collapse to a single point, as all tasks can potentially update all domain-specific centers. When such implicit clustering occurs (Figure \ref{fig:cluster} (c)), the benefits of the added contrastiveness are less significant. We also compare with supervised versions of MMAML and TSA-MAML, where we assume access to each task's source. For MMAML we include an extra cross-entropy loss term for the task encoder in conjunction with an additional classification head (similar to tri-M). For supervised TSA-MAML we train the router with supervised contrastive learning~\cite{khosla2020supervised}. As seen in Tables \ref{table:mmaml},\ref{table:tsamaml}, the results with domain labels are better than their unsupervised counterparts, and only slightly better than our contrastive method. This indicates that our approach gets close to what can be perceived as optimal in these experimental settings.

\begin{figure}[t]
\begin{center}
\centerline{\includegraphics[width=0.9\linewidth]{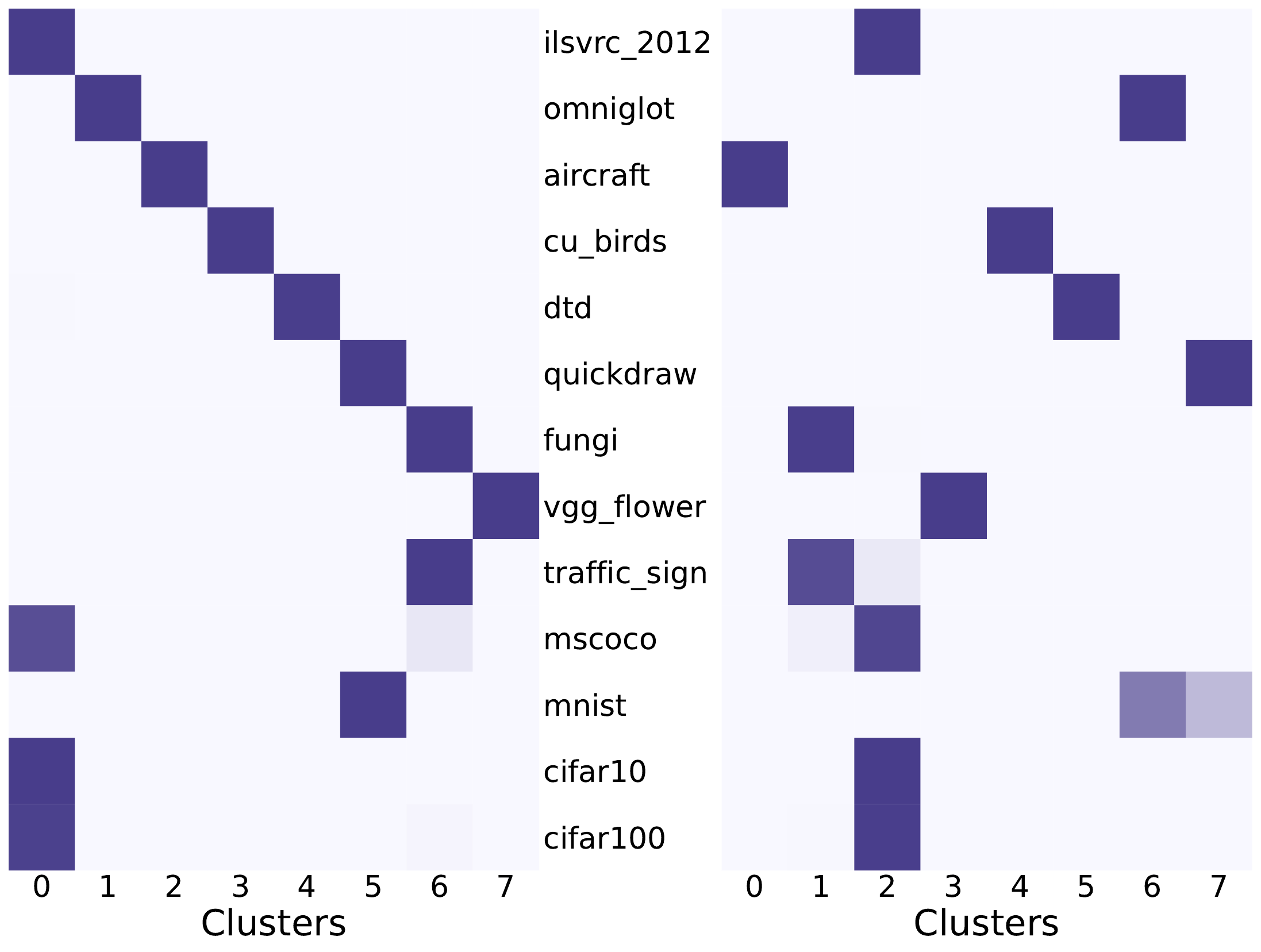}}
\caption{Cluster/domain heatmap. We plot the cluster assignment of tasks for Tri-M (left) and our con-Tri-M (right). Even in the absence of domain labels (right) we are able to get assignements that closely resemble the ones obtained in a supervised manner (left)}
\label{fig:trim}
\end{center}
\vspace{-0.5in}
\end{figure}

Finally, for Tri-M, we see in Table \ref{table:trim} that we get no deterioration in performance even after eliminating the need for task domain labels. In Figure~\ref{fig:trim} we plot the cluster/domain heatmap and we see that even in our unsupervised setting, there are distinct cluster-domain pairs being formed due to our contrastive framework.

\begin{table}[t!]
\caption{The performance (accuracy) for the Tri-M and its contrastive counterpart. We see that our contrastive approach perfroms just as well, \textbf{without the need for domain labels.}}
\small
\vspace{-0.2in}
\label{sample-table}
\begin{center}
\begin{tabular}{lll}
\hline
                               & \textbf{Tri-M} & \textbf{con-Tri-M} \\ \hline
\textbf{ImageNet}              & $54.5\pm1.1$   & $54.3\pm1.1$       \\
\textbf{Omniglot}              & $92.2\pm0.6$   & $92.0\pm0.5$       \\
\textbf{Aircraft}              & $81.8\pm0.6$   & $82.8\pm0.7$       \\
\textbf{Birds}                 & $75.9\pm0.9$   & $75.6\pm0.9$       \\
\textbf{Textures}              & $69.5\pm0.8$   & $70.2\pm0.7$       \\
\textbf{QuickDraw}             & $76.9\pm 0.8$  & $77.3\pm0.8$       \\
\textbf{Fungi}                 & $48.1\pm1.2$   & $47.1\pm1.2$       \\
\textbf{VGGFlower}             & $91.5\pm0.5$   & $90.5\pm0.6$       \\ \hline
\textbf{TrafficSigns}          & $57.5\pm1.1$   & $58.5\pm1.1$       \\
\textbf{MSCOCO}                & $49.7\pm1.1$   & $49.8\pm1.1$       \\
\textbf{MNIST}                 & $94.3\pm0.4$   & $95.0\pm0.4$       \\
\textbf{CIFAR10}               & $74.0\pm0.8$   & $73.1\pm0.8$       \\
\textbf{CIFAR100}              & $62.3\pm1.0$   & $62.1\pm1.0$       \\ \hline
\textbf{In domain Average}     & $73.8$         & $73.7$             \\
\textbf{Out of domain Average} & $67.6$         & $67.7$             \\
\textbf{Overall Average}       & $71.4 $        & $71.4$             \\ \hline
\end{tabular}
\end{center}
\label{table:trim}
\vspace{-0.3in}
\end{table}

%Concluding, we need to stress that the only hyperparameter that our framework is sensitive to is the mixing coefficient $\lambda$ which mixes the contrastive loss with the other losses present. Other than that, our method can be applied in the exact same way to other popular methods in the literature. Furthermore, we believe that the significance of this approach would become more obvious in cases where the different domains have some overlap, which consequently would further hinder non-contrastive methods.

\subsection{Computational Benefits}
Our task-level contrastiveness can also reduce the memory and time requirements of algorithms. Here, we highlight these benefits for two cases: the aforementioned parameter-efficient TSA-MAML variants and Hierarchical Distillation in URL~\cite{94ba0ea9d86744f69497ae97aa68c584}. URL involves aligning the representations of the universal network with multiple single-task networks, a process that can become memory and compute-intensive as the number of tasks (T) increases. By dividing all tasks into different task groups, it is possible to learn a single network for each group $g$ by distilling the single-task representations of the tasks within that group. The universal network can then be obtained by distilling these group networks. The construction of these groups can be done randomly or by utilizing contrastive task representations (see Table~\ref{table:augs}) and performing K-means to obtain centers, which will be used to classify tasks into each group. 
Groups therefore are created by taking into account a notion of task similarity.
This leads to lower memory requirements and slightly better final average accuracy(see Table~\ref{table:compbenefits}).

\section{Conclusion}
\label{conclusion}
We introduced task-level contrastiveness to enhance meta-learning and few-shot learning performance in heterogeneous task settings. Our tool seamlessly integrates with most existing methods, demonstrating improvements in generalization, reduced computational and memory requirements, and the elimination of certain modeling assumptions. By selecting key algorithms from the literature, we showed how our approach effectively enhances and enables domain-aware modulation and unsupervised domain-expert routing—the two mechanisms relied upon by state of the art methods. In future work, we plan to explore various settings, such as few-shot dense prediction~\cite{kim2023universal}, which represents another interesting application of few-shot learning. Additionally, multi-task settings or mixture of experts approaches could potentially benefit from our task-level contrastiveness. One major limitation of our approach is our choice of task augmentations. While mixing is a powerful technique, several studies have noted that a support/query split, which is necessary for our method, is not always required for meta-learning algorithms~\cite{bai2021important}. Exploring alternatives to our current augmentation strategies could further enhance the applicability and effectiveness of our approach.

{
    \small
    \bibliographystyle{ieeenat_fullname}
    \bibliography{main}

\begin{thebibliography}{61}
\providecommand{\natexlab}[1]{#1}
\providecommand{\url}[1]{\texttt{#1}}
\expandafter\ifx\csname urlstyle\endcsname\relax
  \providecommand{\doi}[1]{doi: #1}\else
  \providecommand{\doi}{doi: \begingroup \urlstyle{rm}\Url}\fi

\bibitem[Bai et~al.(2021)Bai, Chen, Zhou, Zhao, Lee, Kakade, Wang, and Xiong]{bai2021important}
Yu Bai, Minshuo Chen, Pan Zhou, Tuo Zhao, Jason Lee, Sham Kakade, Huan Wang, and Caiming Xiong.
\newblock How important is the train-validation split in meta-learning?
\newblock In \emph{International Conference on Machine Learning}, pages 543--553. PMLR, 2021.

\bibitem[Balestriero et~al.(2023)Balestriero, Ibrahim, Sobal, Morcos, Shekhar, Goldstein, Bordes, Bardes, Mialon, Tian, Schwarzschild, Wilson, Geiping, Garrido, Fernandez, Bar, Pirsiavash, LeCun, and Goldblum]{balestriero2023cookbook}
Randall Balestriero, Mark Ibrahim, Vlad Sobal, Ari Morcos, Shashank Shekhar, Tom Goldstein, Florian Bordes, Adrien Bardes, Gregoire Mialon, Yuandong Tian, Avi Schwarzschild, Andrew~Gordon Wilson, Jonas Geiping, Quentin Garrido, Pierre Fernandez, Amir Bar, Hamed Pirsiavash, Yann LeCun, and Micah Goldblum.
\newblock A cookbook of self-supervised learning, 2023.

\bibitem[Bateni et~al.(2020)Bateni, Goyal, Masrani, Wood, and Sigal]{Bateni_2020_CVPR}
Peyman Bateni, Raghav Goyal, Vaden Masrani, Frank Wood, and Leonid Sigal.
\newblock Improved few-shot visual classification.
\newblock In \emph{Proceedings of the IEEE/CVF Conference on Computer Vision and Pattern Recognition (CVPR)}, 2020.

\bibitem[Bateni et~al.(2022)Bateni, Barber, Van~de Meent, and Wood]{bateni2022enhancing}
Peyman Bateni, Jarred Barber, Jan-Willem Van~de Meent, and Frank Wood.
\newblock Enhancing few-shot image classification with unlabelled examples.
\newblock In \emph{Proceedings of the IEEE/CVF Winter Conference on Applications of Computer Vision}, pages 2796--2805, 2022.

\bibitem[Ben~Zaken et~al.(2022)Ben~Zaken, Goldberg, and Ravfogel]{ben-zaken-etal-2022-bitfit}
Elad Ben~Zaken, Yoav Goldberg, and Shauli Ravfogel.
\newblock {B}it{F}it: Simple parameter-efficient fine-tuning for transformer-based masked language-models.
\newblock In \emph{Proceedings of the 60th Annual Meeting of the Association for Computational Linguistics (Volume 2: Short Papers)}, pages 1--9, Dublin, Ireland, 2022. Association for Computational Linguistics.

\bibitem[Bertinetto et~al.(2019)Bertinetto, Henriques, Torr, and Vedaldi]{bertinetto2019metalearning}
L Bertinetto, J Henriques, P Torr, and A Vedaldi.
\newblock Meta-learning with differentiable closed-form solvers.
\newblock In \emph{International Conference on Learning Representations (ICLR), 2019}. International Conference on Learning Representations, 2019.

\bibitem[Cai et~al.(2020)Cai, Gan, Massachusetts Institute~of Technology, and Massachusetts Institute~of Technology]{10.5555/3495724.3496671}
Han Cai, Chuang Gan, Ligeng~Zhu Massachusetts Institute~of Technology, and Song~Han Massachusetts Institute~of Technology.
\newblock Tinytl: reduce memory, not parameters for efficient on-device learning.
\newblock In \emph{Proceedings of the 34th International Conference on Neural Information Processing Systems}, Red Hook, NY, USA, 2020. Curran Associates Inc.

\bibitem[Caron et~al.(2018)Caron, Bojanowski, Joulin, and Douze]{caron2018deep}
Mathilde Caron, Piotr Bojanowski, Armand Joulin, and Matthijs Douze.
\newblock Deep clustering for unsupervised learning of visual features.
\newblock In \emph{Proceedings of the European conference on computer vision (ECCV)}, pages 132--149, 2018.

\bibitem[Chen et~al.(2020)Chen, Kornblith, Norouzi, and Hinton]{chen2020simple}
Ting Chen, Simon Kornblith, Mohammad Norouzi, and Geoffrey Hinton.
\newblock A simple framework for contrastive learning of visual representations.
\newblock In \emph{International conference on machine learning}, pages 1597--1607. PMLR, 2020.

\bibitem[Chuang et~al.(2020)Chuang, Robinson, Lin, Torralba, and Jegelka]{NEURIPS2020_63c3ddcc}
Ching-Yao Chuang, Joshua Robinson, Yen-Chen Lin, Antonio Torralba, and Stefanie Jegelka.
\newblock Debiased contrastive learning.
\newblock In \emph{Advances in Neural Information Processing Systems}, pages 8765--8775. Curran Associates, Inc., 2020.

\bibitem[Dosovitskiy et~al.(2021)Dosovitskiy, Beyer, Kolesnikov, Weissenborn, Zhai, Unterthiner, Dehghani, Minderer, Heigold, Gelly, Uszkoreit, and Houlsby]{vit}
Alexey Dosovitskiy, Lucas Beyer, Alexander Kolesnikov, Dirk Weissenborn, Xiaohua Zhai, Thomas Unterthiner, Mostafa Dehghani, Matthias Minderer, Georg Heigold, Sylvain Gelly, Jakob Uszkoreit, and Neil Houlsby.
\newblock An image is worth 16x16 words: Transformers for image recognition at scale.
\newblock In \emph{9th International Conference on Learning Representations, {ICLR} 2021, Virtual Event, Austria, May 3-7, 2021}. OpenReview.net, 2021.

\bibitem[Dvornik et~al.(2020)Dvornik, Schmid, and Mairal]{dvornik2020selecting}
Nikita Dvornik, Cordelia Schmid, and Julien Mairal.
\newblock Selecting relevant features from a multi-domain representation for few-shot classification.
\newblock In \emph{Computer Vision – ECCV 2020: 16th European Conference, Glasgow, UK, August 23–28, 2020, Proceedings, Part X}, page 769–786, Berlin, Heidelberg, 2020. Springer-Verlag.

\bibitem[Finn et~al.(2017)Finn, Abbeel, and Levine]{finn2017modelagnostic}
Chelsea Finn, Pieter Abbeel, and Sergey Levine.
\newblock Model-agnostic meta-learning for fast adaptation of deep networks.
\newblock In \emph{Proceedings of the 34th International Conference on Machine Learning}, pages 1126--1135. PMLR, 2017.

\bibitem[Garnelo et~al.(2018)Garnelo, Rosenbaum, Maddison, Ramalho, Saxton, Shanahan, Teh, Rezende, and Eslami]{garnelo2018conditional}
Marta Garnelo, Dan Rosenbaum, Christopher Maddison, Tiago Ramalho, David Saxton, Murray Shanahan, Yee~Whye Teh, Danilo Rezende, and SM~Ali Eslami.
\newblock Conditional neural processes.
\newblock In \emph{International conference on machine learning}, pages 1704--1713. PMLR, 2018.

\bibitem[Gondal et~al.(2021)Gondal, Joshi, Rahaman, Bauer, Wuthrich, and Sch{\"o}lkopf]{gondal2021function}
Muhammad~Waleed Gondal, Shruti Joshi, Nasim Rahaman, Stefan Bauer, Manuel Wuthrich, and Bernhard Sch{\"o}lkopf.
\newblock Function contrastive learning of transferable meta-representations.
\newblock In \emph{International Conference on Machine Learning}, pages 3755--3765. PMLR, 2021.

\bibitem[Gordon et~al.(2019)Gordon, Bruinsma, Foong, Requeima, Dubois, and Turner]{gordon2020convolutional}
Jonathan Gordon, Wessel~P Bruinsma, Andrew~YK Foong, James Requeima, Yann Dubois, and Richard~E Turner.
\newblock Convolutional conditional neural processes.
\newblock In \emph{International Conference on Learning Representations}, 2019.

\bibitem[He et~al.(2020)He, Fan, Wu, Xie, and Girshick]{he2020momentum}
Kaiming He, Haoqi Fan, Yuxin Wu, Saining Xie, and Ross Girshick.
\newblock Momentum contrast for unsupervised visual representation learning.
\newblock In \emph{Proceedings of the IEEE/CVF conference on computer vision and pattern recognition}, pages 9729--9738, 2020.

\bibitem[Hjelm et~al.(2018)Hjelm, Fedorov, Lavoie-Marchildon, Grewal, Bachman, Trischler, and Bengio]{hjelm2019learning}
R~Devon Hjelm, Alex Fedorov, Samuel Lavoie-Marchildon, Karan Grewal, Phil Bachman, Adam Trischler, and Yoshua Bengio.
\newblock Learning deep representations by mutual information estimation and maximization.
\newblock In \emph{International Conference on Learning Representations}, 2018.

\bibitem[Hu et~al.(2022)Hu, Li, St{\"u}hmer, Kim, and Hospedales]{hu2022pushing}
Shell~Xu Hu, Da Li, Jan St{\"u}hmer, Minyoung Kim, and Timothy~M Hospedales.
\newblock Pushing the limits of simple pipelines for few-shot learning: External data and fine-tuning make a difference.
\newblock In \emph{Proceedings of the IEEE/CVF Conference on Computer Vision and Pattern Recognition}, pages 9068--9077, 2022.

\bibitem[Jang et~al.(2023)Jang, Lee, and Shin]{jang2023unsupervised}
Huiwon Jang, Hankook Lee, and Jinwoo Shin.
\newblock Unsupervised meta-learning via few-shot pseudo-supervised contrastive learning.
\newblock \emph{arXiv preprint arXiv:2303.00996}, 2023.

\bibitem[Jiang et~al.(2022)Jiang, Kwok, and Zhang]{pmlr-v162-jiang22b}
Weisen Jiang, James Kwok, and Yu Zhang.
\newblock Subspace learning for effective meta-learning.
\newblock In \emph{Proceedings of the 39th International Conference on Machine Learning}, pages 10177--10194. PMLR, 2022.

\bibitem[Khosla et~al.(2020)Khosla, Teterwak, Wang, Sarna, Tian, Isola, Maschinot, Liu, and Krishnan]{khosla2020supervised}
Prannay Khosla, Piotr Teterwak, Chen Wang, Aaron Sarna, Yonglong Tian, Phillip Isola, Aaron Maschinot, Ce Liu, and Dilip Krishnan.
\newblock Supervised contrastive learning.
\newblock \emph{Advances in neural information processing systems}, 33:\penalty0 18661--18673, 2020.

\bibitem[Kim et~al.(2023)Kim, Kim, Cho, Luo, and Hong]{kim2023universal}
Donggyun Kim, Jinwoo Kim, Seongwoong Cho, Chong Luo, and Seunghoon Hong.
\newblock Universal few-shot learning of dense prediction tasks with visual token matching.
\newblock \emph{arXiv preprint arXiv:2303.14969}, 2023.

\bibitem[Krizhevsky et~al.(2012)Krizhevsky, Sutskever, and Hinton]{NIPS2012_c399862d}
Alex Krizhevsky, Ilya Sutskever, and Geoffrey~E Hinton.
\newblock Imagenet classification with deep convolutional neural networks.
\newblock In \emph{Advances in Neural Information Processing Systems}. Curran Associates, Inc., 2012.

\bibitem[Lee and Whan~Yoon(2024)]{pmlr-v238-lee24b}
Jae-Jun Lee and Sung Whan~Yoon.
\newblock {XB-MAML}: Learning expandable basis parameters for effective meta-learning with wide task coverage.
\newblock In \emph{Proceedings of The 27th International Conference on Artificial Intelligence and Statistics}, pages 3196--3204. PMLR, 2024.

\bibitem[Lee et~al.(2019)Lee, Maji, Ravichandran, and Soatto]{lee2019metalearning}
Kwonjoon Lee, Subhransu Maji, Avinash Ravichandran, and Stefano Soatto.
\newblock Meta-learning with differentiable convex optimization.
\newblock In \emph{Proceedings of the IEEE/CVF conference on computer vision and pattern recognition}, pages 10657--10665, 2019.

\bibitem[Li et~al.(2021{\natexlab{a}})Li, Liu, and Bilen]{li2021universal}
Wei-Hong Li, Xialei Liu, and Hakan Bilen.
\newblock Universal representation learning from multiple domains for few-shot classification.
\newblock In \emph{Proceedings of the IEEE/CVF International Conference on Computer Vision}, pages 9526--9535, 2021{\natexlab{a}}.

\bibitem[Li et~al.(2022)Li, Liu, and Bilen]{li2022crossdomain}
Wei-Hong Li, Xialei Liu, and Hakan Bilen.
\newblock Cross-domain few-shot learning with task-specific adapters.
\newblock In \emph{Proceedings of the IEEE/CVF Conference on Computer Vision and Pattern Recognition}, pages 7161--7170, 2022.

\bibitem[Li et~al.(2023)Li, Liu, and Bilen]{94ba0ea9d86744f69497ae97aa68c584}
Wei-Hong Li, Xialei Liu, and Hakan Bilen.
\newblock Universal representations: A unified look at multiple task and domain learning.
\newblock \emph{International Journal of Computer Vision}, pages 1--25, 2023.
\newblock This work was partly supported by the EPSRC programme grant Visual AI EP/T028572/1 and a Huawei Technologies R\&D (UK) project.

\bibitem[Li et~al.(2021{\natexlab{b}})Li, Hu, Liu, Peng, Zhou, and Peng]{li2020contrastive}
Yunfan Li, Peng Hu, Zitao Liu, Dezhong Peng, Joey~Tianyi Zhou, and Xi Peng.
\newblock Contrastive clustering.
\newblock In \emph{Proceedings of the AAAI conference on artificial intelligence}, pages 8547--8555, 2021{\natexlab{b}}.

\bibitem[Liu et~al.(2021)Liu, Lee, Zhu, Chen, Shi, and Yang]{9710504}
Yanbin Liu, Juho Lee, Linchao Zhu, Ling Chen, Humphrey Shi, and Yi Yang.
\newblock A multi-mode modulator for multi-domain few-shot classification.
\newblock In \emph{2021 IEEE/CVF International Conference on Computer Vision (ICCV)}, pages 8433--8442, 2021.

\bibitem[Ni et~al.(2021)Ni, Goldblum, Sharaf, Kong, and Goldstein]{ni2021data}
Renkun Ni, Micah Goldblum, Amr Sharaf, Kezhi Kong, and Tom Goldstein.
\newblock Data augmentation for meta-learning.
\newblock In \emph{International Conference on Machine Learning}, pages 8152--8161. PMLR, 2021.

\bibitem[Oquab et~al.(2023)Oquab, Darcet, Moutakanni, Vo, Szafraniec, Khalidov, Fernandez, Haziza, Massa, El-Nouby, Howes, Huang, Xu, Sharma, Li, Galuba, Rabbat, Assran, Ballas, Synnaeve, Misra, Jegou, Mairal, Labatut, Joulin, and Bojanowski]{oquab2023dinov2}
Maxime Oquab, Timothée Darcet, Theo Moutakanni, Huy~V. Vo, Marc Szafraniec, Vasil Khalidov, Pierre Fernandez, Daniel Haziza, Francisco Massa, Alaaeldin El-Nouby, Russell Howes, Po-Yao Huang, Hu Xu, Vasu Sharma, Shang-Wen Li, Wojciech Galuba, Mike Rabbat, Mido Assran, Nicolas Ballas, Gabriel Synnaeve, Ishan Misra, Herve Jegou, Julien Mairal, Patrick Labatut, Armand Joulin, and Piotr Bojanowski.
\newblock Dinov2: Learning robust visual features without supervision, 2023.

\bibitem[Oreshkin et~al.(2018)Oreshkin, Rodriguez, and Lacoste]{oreshkin2019tadam}
Boris~N. Oreshkin, Pau Rodriguez, and Alexandre Lacoste.
\newblock Tadam: task dependent adaptive metric for improved few-shot learning.
\newblock In \emph{Proceedings of the 32nd International Conference on Neural Information Processing Systems}, page 719–729, Red Hook, NY, USA, 2018. Curran Associates Inc.

\bibitem[Peng and Pan(2023)]{10061493}
Danni Peng and Sinno~Jialin Pan.
\newblock Clustered task-aware meta-learning by learning from learning paths.
\newblock \emph{IEEE Transactions on Pattern Analysis and Machine Intelligence}, 45\penalty0 (8):\penalty0 9426--9438, 2023.

\bibitem[Perera and Halgamuge(2024)]{perera2024discriminative}
Rashindrie Perera and Saman Halgamuge.
\newblock Discriminative sample-guided and parameter-efficient feature space adaptation for cross-domain few-shot learning.
\newblock In \emph{Proceedings of the IEEE/CVF Conference on Computer Vision and Pattern Recognition}, pages 23794--23804, 2024.

\bibitem[Perez et~al.(2018)Perez, Strub, De~Vries, Dumoulin, and Courville]{perez2017film}
Ethan Perez, Florian Strub, Harm De~Vries, Vincent Dumoulin, and Aaron Courville.
\newblock Film: Visual reasoning with a general conditioning layer.
\newblock In \emph{Proceedings of the AAAI conference on artificial intelligence}, 2018.

\bibitem[Poulakakis-Daktylidis and Jamali-Rad(2024)]{poulakakis-daktylidis2024beclr}
Stylianos Poulakakis-Daktylidis and Hadi Jamali-Rad.
\newblock {BECLR}: Batch enhanced contrastive unsupervised few-shot learning.
\newblock In \emph{The Twelfth International Conference on Learning Representations}, 2024.

\bibitem[Radford et~al.(2021)Radford, Kim, Hallacy, Ramesh, Goh, Agarwal, Sastry, Askell, Mishkin, Clark, et~al.]{radford2021learning}
Alec Radford, Jong~Wook Kim, Chris Hallacy, Aditya Ramesh, Gabriel Goh, Sandhini Agarwal, Girish Sastry, Amanda Askell, Pamela Mishkin, Jack Clark, et~al.
\newblock Learning transferable visual models from natural language supervision.
\newblock In \emph{International conference on machine learning}, pages 8748--8763. PMLR, 2021.

\bibitem[Raghu et~al.(2019)Raghu, Raghu, Bengio, and Vinyals]{raghu2020rapid}
Aniruddh Raghu, Maithra Raghu, Samy Bengio, and Oriol Vinyals.
\newblock Rapid learning or feature reuse? towards understanding the effectiveness of maml.
\newblock In \emph{International Conference on Learning Representations}, 2019.

\bibitem[Ravi and Larochelle(2017)]{Sachin2017}
Sachin Ravi and Hugo Larochelle.
\newblock Optimization as a model for few-shot learning.
\newblock In \emph{In International Conference on Learning Representations (ICLR)}, 2017.

\bibitem[Requeima et~al.(2019)Requeima, Gordon, Bronskill, Nowozin, and Turner]{requeima2020fast}
James Requeima, Jonathan Gordon, John Bronskill, Sebastian Nowozin, and Richard~E Turner.
\newblock Fast and flexible multi-task classification using conditional neural adaptive processes.
\newblock \emph{Advances in Neural Information Processing Systems}, 32, 2019.

\bibitem[Robinson et~al.(2021)Robinson, Chuang, Sra, and Jegelka]{robinson2021contrastive}
Joshua Robinson, Ching-Yao Chuang, Suvrit Sra, and Stefanie Jegelka.
\newblock Contrastive learning with hard negative samples.
\newblock In \emph{International Conference on Learning Representations (ICLR)}, 2021.

\bibitem[Shidani et~al.(2024)Shidani, Hjelm, Ramapuram, Webb, Dhekane, and Busbridge]{shidani2024poly}
Amitis Shidani, Devon Hjelm, Jason Ramapuram, Russ Webb, Eeshan~Gunesh Dhekane, and Dan Busbridge.
\newblock Poly-view contrastive learning.
\newblock \emph{arXiv preprint arXiv:2403.05490}, 2024.

\bibitem[Sinha et~al.(2021)Sinha, Ayush, Song, Uzkent, Jin, and Ermon]{sinha2021negative}
Abhishek Sinha, Kumar Ayush, Jiaming Song, Burak Uzkent, Hongxia Jin, and Stefano Ermon.
\newblock Negative data augmentation.
\newblock In \emph{International Conference on Learning Representations}, 2021.

\bibitem[Snell et~al.(2017)Snell, Swersky, and Zemel]{snell2017prototypical}
Jake Snell, Kevin Swersky, and Richard Zemel.
\newblock Prototypical networks for few-shot learning.
\newblock In \emph{Proceedings of the 31st International Conference on Neural Information Processing Systems}, page 4080–4090, Red Hook, NY, USA, 2017. Curran Associates Inc.

\bibitem[Tian et~al.(2020)Tian, Sun, Poole, Krishnan, Schmid, and Isola]{tian2020makes}
Yonglong Tian, Chen Sun, Ben Poole, Dilip Krishnan, Cordelia Schmid, and Phillip Isola.
\newblock What makes for good views for contrastive learning?
\newblock In \emph{Proceedings of the 34th International Conference on Neural Information Processing Systems}, Red Hook, NY, USA, 2020. Curran Associates Inc.

\bibitem[Triantafillou et~al.(2020)Triantafillou, Zhu, Dumoulin, Lamblin, Evci, Xu, Goroshin, Gelada, Swersky, Manzagol, and Larochelle]{triantafillou2020metadataset}
Eleni Triantafillou, Tyler Zhu, Vincent Dumoulin, Pascal Lamblin, Utku Evci, Kelvin Xu, Ross Goroshin, Carles Gelada, Kevin~Jordan Swersky, Pierre-Antoine Manzagol, and Hugo Larochelle.
\newblock Meta-dataset: A dataset of datasets for learning to learn from few examples.
\newblock In \emph{International Conference on Learning Representations (submission)}, 2020.

\bibitem[Triantafillou et~al.(2021)Triantafillou, Larochelle, Zemel, and Dumoulin]{triantafillou2021learning}
Eleni Triantafillou, Hugo Larochelle, Richard Zemel, and Vincent Dumoulin.
\newblock Learning a universal template for few-shot dataset generalization.
\newblock In \emph{International Conference on Machine Learning}, pages 10424--10433. PMLR, 2021.

\bibitem[van~den Oord et~al.(2019)van~den Oord, Li, and Vinyals]{oord2019representation}
Aaron van~den Oord, Yazhe Li, and Oriol Vinyals.
\newblock Representation learning with contrastive predictive coding, 2019.

\bibitem[Vettoruzzo et~al.(2023)Vettoruzzo, Bouguelia, and Rögnvaldsson]{10191944}
Anna Vettoruzzo, Mohamed-Rafik Bouguelia, and Thorsteinn Rögnvaldsson.
\newblock Meta-learning from multimodal task distributions using multiple sets of meta-parameters.
\newblock In \emph{2023 International Joint Conference on Neural Networks (IJCNN)}, pages 1--8, 2023.

\bibitem[Vinyals et~al.(2017)Vinyals, Blundell, Lillicrap, Kavukcuoglu, and Wierstra]{vinyals2017matching}
Oriol Vinyals, Charles Blundell, Timothy Lillicrap, Koray Kavukcuoglu, and Daan Wierstra.
\newblock Matching networks for one shot learning, 2017.

\bibitem[Vuorio et~al.(2019)Vuorio, Sun, Hu, and Lim]{vuorio2019multimodal}
Risto Vuorio, Shao-Hua Sun, Hexiang Hu, and Joseph~J Lim.
\newblock Multimodal model-agnostic meta-learning via task-aware modulation.
\newblock \emph{Advances in neural information processing systems}, 32, 2019.

\bibitem[Wu et~al.(2018)Wu, Xiong, Yu, and Lin]{wu2018unsupervised}
Zhirong Wu, Yuanjun Xiong, Stella~X Yu, and Dahua Lin.
\newblock Unsupervised feature learning via non-parametric instance discrimination.
\newblock In \emph{Proceedings of the IEEE conference on computer vision and pattern recognition}, pages 3733--3742, 2018.

\bibitem[Xu(2023)]{Xu2023ExploringEF}
C. Xu.
\newblock Exploring efficient few-shot adaptation for vision transformers.
\newblock \emph{Trans. Mach. Learn. Res.}, 2022, 2023.

\bibitem[Yang et~al.(2022)Yang, Wang, and Zhu]{yang2022few}
Zhanyuan Yang, Jinghua Wang, and Yingying Zhu.
\newblock Few-shot classification with contrastive learning.
\newblock In \emph{European conference on computer vision}, pages 293--309. Springer, 2022.

\bibitem[Yao et~al.(2019{\natexlab{a}})Yao, Wei, Huang, and Li]{yao2019hierarchically}
Huaxiu Yao, Ying Wei, Junzhou Huang, and Zhenhui Li.
\newblock Hierarchically structured meta-learning.
\newblock In \emph{International Conference on Machine Learning}, pages 7045--7054. PMLR, 2019{\natexlab{a}}.

\bibitem[Yao et~al.(2019{\natexlab{b}})Yao, Wu, Tao, Li, Ding, Li, and Li]{yao2020automated}
Huaxiu Yao, Xian Wu, Zhiqiang Tao, Yaliang Li, Bolin Ding, Ruirui Li, and Zhenhui Li.
\newblock Automated relational meta-learning.
\newblock In \emph{International Conference on Learning Representations}, 2019{\natexlab{b}}.

\bibitem[Yeh et~al.(2022)Yeh, Hong, Hsu, Liu, Chen, and LeCun]{yeh2022decoupled}
Chun-Hsiao Yeh, Cheng-Yao Hong, Yen-Chi Hsu, Tyng-Luh Liu, Yubei Chen, and Yann LeCun.
\newblock Decoupled contrastive learning.
\newblock In \emph{European Conference on Computer Vision}, pages 668--684. Springer, 2022.

\bibitem[Zaheer et~al.(2017)Zaheer, Kottur, Ravanbhakhsh, P\'{o}czos, Salakhutdinov, and Smola]{zaheer2018deep}
Manzil Zaheer, Satwik Kottur, Siamak Ravanbhakhsh, Barnab\'{a}s P\'{o}czos, Ruslan Salakhutdinov, and Alexander~J Smola.
\newblock Deep sets.
\newblock In \emph{Proceedings of the 31st International Conference on Neural Information Processing Systems}, page 3394–3404, Red Hook, NY, USA, 2017. Curran Associates Inc.

\bibitem[Zhou et~al.(2020)Zhou, Zou, Yuan, Feng, Xiong, and Hoi]{zhou2020task}
Pan Zhou, Yingtian Zou, Xiaotong Yuan, Jiashi Feng, Caiming Xiong, and SC Hoi.
\newblock Task similarity aware meta learning: Theory-inspired improvement on maml.
\newblock In \emph{4th Workshop on Meta-Learning at NeurIPS}, 2020.

\end{thebibliography}
}
%\input{cvpr2025/sec/X_suppl2}


\end{document}